\documentclass[letterpaper]{article} 
\usepackage{aaai2026}
\usepackage{times}  
\usepackage{helvet}  
\usepackage{courier}  
\usepackage[hyphens]{url}  
\usepackage{graphicx} 
\usepackage{natbib}  
\usepackage{caption} 
\usepackage{algorithm}
\usepackage{algorithmic}

\usepackage{amsmath}
\usepackage{amssymb}
\usepackage{cleveref}

\usepackage{indentfirst}
\usepackage{xcolor}
\usepackage{multirow}
\usepackage{booktabs}
\usepackage{adjustbox}
\usepackage{makecell}
\usepackage{utfsym}
\usepackage{xcolor}
\usepackage[table]{xcolor}
\usepackage{colortbl}
\usepackage{tabularray}
\usepackage{makecell}

\UseTblrLibrary{booktabs}

\usepackage{newfloat}
\usepackage{listings}
\DeclareCaptionStyle{ruled}{labelfont=normalfont,labelsep=colon,strut=off} 
\floatstyle{ruled}
\newfloat{listing}{tb}{lst}{}
\floatname{listing}{Listing}
\title{MambaOVSR: Multiscale Fusion with Global Motion Modeling for Chinese Opera Video Super-Resolution}
\author {
    Hua Chang,
    Xin Xu,
    Wei Liu,
    Wei Wang,
    Xin Yuan,
    Kui Jiang
}

\usepackage{bibentry}

\begin{document}

\maketitle

\begin{abstract}
Chinese opera is celebrated for preserving classical art. However, early filming equipment limitations have degraded videos of last-century performances by renowned artists (\emph{e.g.}, low frame rates and resolution), hindering archival efforts. Although space-time video super-resolution (STVSR) has advanced significantly, applying it directly to opera videos remains challenging. The scarcity of datasets impedes the recovery of high-frequency details, and existing STVSR methods lack global modeling capabilities—compromising visual quality when handling opera’s characteristic large motions. To address these challenges, we pioneer a large-scale Chinese Opera Video Clip (COVC) dataset and propose the \textbf{Mamba}-based multiscale fusion network for space-time \textbf{O}pera \textbf{V}ideo \textbf{S}uper-\textbf{R}esolution (MambaOVSR). Specifically, MambaOVSR involves three novel components: the Global Fusion Module (GFM) for motion modeling through a multiscale alternating scanning mechanism, and the Multiscale Synergistic Mamba Module (MSMM) for alignment across different sequence lengths. Additionally, our MambaVR block resolves feature artifacts and positional information loss during alignment. Experimental results on the COVC dataset show that MambaOVSR significantly outperforms the SOTA STVSR method by an average of 1.86 dB in terms of PSNR. \textit{Dataset and Code will be publicly released}.

\end{abstract}


\section{Introduction}

\begin{figure}[t]
  \centering
   \includegraphics[width=1\linewidth]{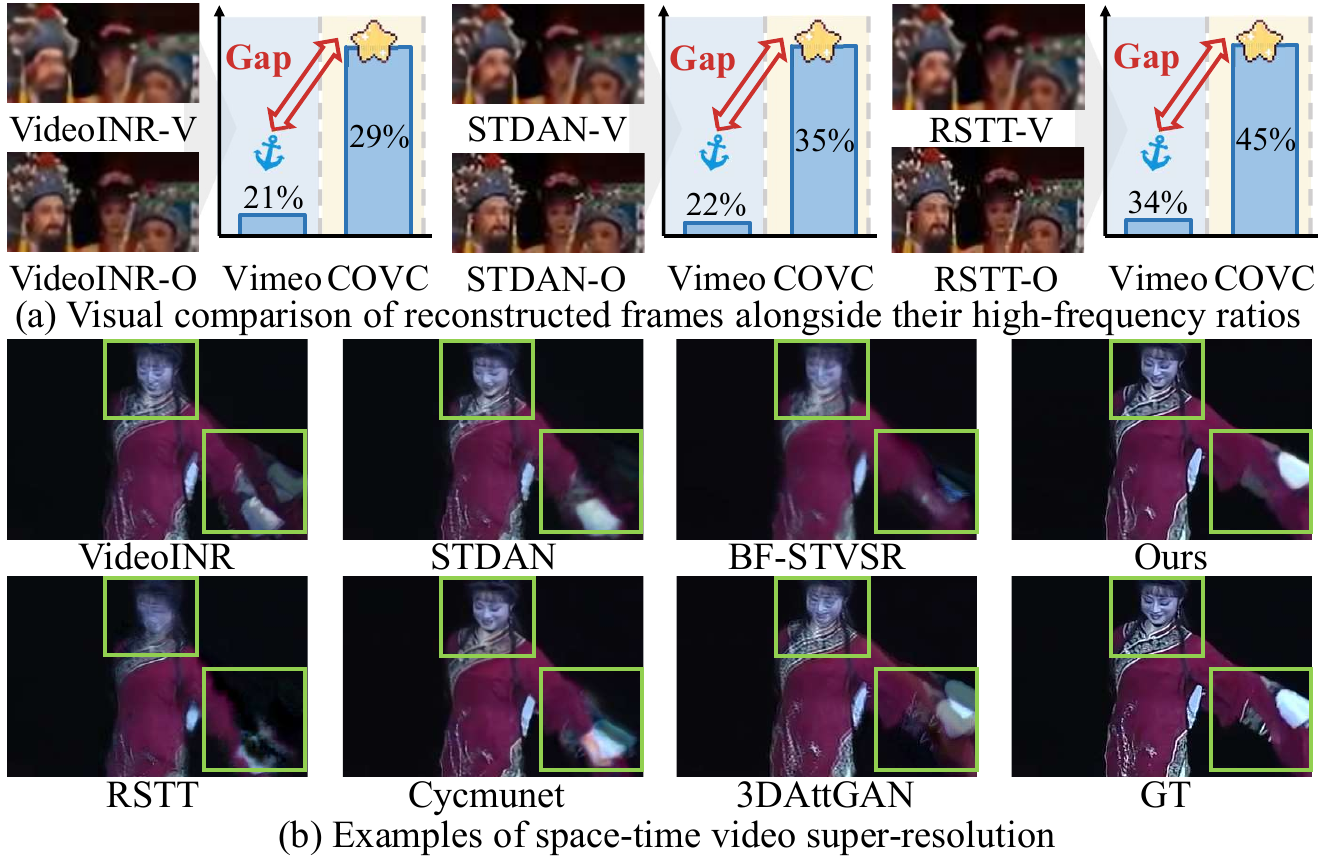}\vspace*{-2mm}
   \caption{(a) Visual comparison and high-frequency content ratios for the same model trained on Vimeo90K (‑V) and COVC (‑O); other methods’ ratios are reported in Appendix Section A. The COVC‑trained model recovers more high-frequency details. (b) Presents that existing methods synthesize intermediate frames with blurring artifacts.}
   \label{issue}\vspace*{-4mm}
\end{figure}

\noindent Chinese opera represents a distinctive performing art of significant cultural value. However, limitations in early filming technology and storage media degradation have left many classic recordings with low resolution and frame rates, severely hindering preservation efforts and scholarly study~\cite{chung2024utilising}.

Space-Time Video Super-Resolution (STVSR), first proposed in 2020 \cite{xiang2020zooming}, enhances both temporal and spatial video resolution. Early approaches combined Video Frame Interpolation (VFI) \cite{cheng2021multiple,liu2024video} and Video Super-Resolution (VSR) \cite{yi2019progressive,li2023multi} techniques but failed to exploit their intrinsic connections, yielding suboptimal results \cite{haris2020space,hu2023cycmunet+}. Subsequent end-to-end frameworks \cite{xiang2020zooming,xu2021temporal,hu2023cycmunet+,wang2023stdan} and efficiency-focused designs \cite{geng2022rstt,hu2022you,hu2023store,fu20243dattgan} improved STVSR for general scenes. However, these methods remain inadequate for opera due to: the lack of domain-specific datasets and insufficient global modeling capabilities. As shown in \Cref{issue}(a), models trained on general datasets (e.g., Vimeo90K \cite{xue2019video}) fail to recover opera-specific high-frequency textures, highlighting the domain gap.

To tackle the aforementioned challenges, we introduce the Chinese Opera Video Clip (COVC) dataset—the first large-scale collection for opera restoration. COVC contains 33 distinct opera videos processed into training septuples following the Vimeo90K dataset \cite{xue2019video}, yielding 104,138 training samples. When retraining existing STVSR methods on COVC, synthesized frames exhibit blurring artifacts (\Cref{issue}(b)), confirming their inability to model opera's large motions.



In this paper, we propose MambaOVSR, a \textbf{Mamba}-based multiscale fusion network for space-time \textbf{O}pera \textbf{V}ideo \textbf{S}uper-\textbf{R}esolution, which effectively addresses large motion modeling (\Cref{issue}(b)). Specifically, our framework features three innovations: \textbf{Global Fusion Module (GFM)}, \textbf{MambaVR Block} and \textbf{Multiscale Synergistic Mamba Module (MSMM)}. GFM synthesizes intermediate frames by blending forward/backward predictions. Each direction employs a pyramid structure with a Multiscale Alternate Scanning Mechanism (MASM) for global multiscale modeling of adjacent frames, followed by 3D convolutions to extract temporal features from interpolated short sequences. The MambaVR block is designed to resolve feature artifacts and positional information loss in Vision Mamba alignment. MSMM leverages MambaVR blocks for granular motion alignment across varying sequence lengths.




Our contributions are summarized as follows:
\begin{itemize}
\item[$\bullet$] We pioneer a large-scale Chinese Opera Video Clip (COVC) dataset and propose the \textbf{Mamba}-based multiscale fusion network for space-time \textbf{O}pera \textbf{V}ideo \textbf{S}uper-\textbf{R}esolution (MambaOVSR).
\item[$\bullet$] We propose the GFM to perform fine-grained holistic modeling of motion between adjacent frames, accurately synthesizing missing intermediate features. Complementarily, a 3D convolution-based module exploits the temporal feature of neighboring frames for refinement.
\item[$\bullet$] We introduce the MambaVR block for global spatial alignment of multi-frame features. Then, MSMM performs multi-scale alignment on sequences of varying lengths, effectively handling large motions.
\item[$\bullet$] We conduct extensive experiments on both the COVC and general Vimeo90K, demonstrating that the proposed MambaOVSR markedly outperforms existing STVSR methods in both quantitative and qualitative evaluations.
\end{itemize}

\section{Related Work}
\subsection{Space-Time Video Super-Resolution} 



\noindent The Space-Time Video Super-Resolution (STVSR) aims to enhance both the spatial and temporal resolution of videos. Compared to the sequential combined Video Super-Resolution (VSR) and Video Frame Interpolation (VFI) methods \cite{zhou2021video}, the jointly optimized framework has smaller parameters and better results \cite{xiang2020zooming}. STARnet \cite{haris2020space} used high- and low-resolution features to synthesize missing intermediate frames. ZSM \cite{xiang2020zooming} combined deformable convolution with ConvLSTM to propagate frame information. Based on this, TMNet \cite{xu2021temporal} implemented arbitrary time-step frame interpolation. Very recently, Cycmunet \cite{hu2023cycmunet+} and STDAN \cite{wang2023stdan} proposed innovative up-and-down projection units (UPU\&DPU) and deformable feature aggregation (DFA) to achieve frame alignment. Furthermore, to improve the inference speed, RSTT \cite{geng2022rstt} proposed an overall model based on Swin Transformer \cite{Liu_2021_ICCV}.  Although these methods perform well on general scene videos, their performance on opera videos is suboptimal due to richer texture details and larger motions.

\subsection{Visual Mamba}

\noindent Due to its linear complexity and efficient selection mechanism, Mamba \cite{gu2023mamba} has achieved impressive results in natural language processing (NLP). VisionMamba \cite{zhu2024vision} and VMamba \cite{Liu2024vmamba} pioneered the application of Mamba in computer vision by using distinct scanning methods to process images. VideoMamba \cite{li2025videomamba} extended Mamba to video understanding by incorporating spatial and temporal position embedding. Furthermore, Video Mamba Suite \cite{chen2024video} explored the role of Mamba in the four phases of video understanding, highlighting its advantages in video handling. VFIMamba \cite{zhang2024vfimamba} achieves SOTA performance in video frame interpolation (VFI) by modeling adjacent frames through an alternating scanning mechanism (ASM). However, ASM only focuses on global motion and cannot model local motion variations, and we propose the Multiscale Alternate Scanning Mechanism (MASM) to model adjacent frame features. For alignment, the original VideoMamba block \cite{li2025videomamba} is limited by feature artifacts and flexibility. To address this problem, we propose MambaVR for global alignment of frames.



\section{Proposed Method}
\label{sec: method}


\subsection{Chinese Opera Video Clip Dataset}
%
\noindent Low‑quality opera videos hinder the art’s preservation and evolution, and their elaborate costumes, sets, and props produce far richer textures than general benchmarks (e.g., Vimeo90K \cite{xue2019video}), causing existing models to fail in this domain (see \Cref{issue}(a)). To address this, we introduce COVC: a large‑scale Chinese opera video clip dataset.

To ensure dataset quality, we curated 33 high‑quality opera videos by bitrate, resolution, and subjective clarity, then extracted continuous frames—omitting any all‑black boundary frames to avoid invalid PSNR measures. Following the Vimeo90K protocol \cite{xue2019video}, every seven consecutive frames form one clip, yielding 115,548 clips (11,410 for testing; the remainder for training). The test set is stratified by visual quality into High (5,120 clips), Medium (3,150 clips), and Low (3,140 clips) tiers.



\begin{figure}[!t]
  \centering
   \includegraphics[width=1\linewidth]{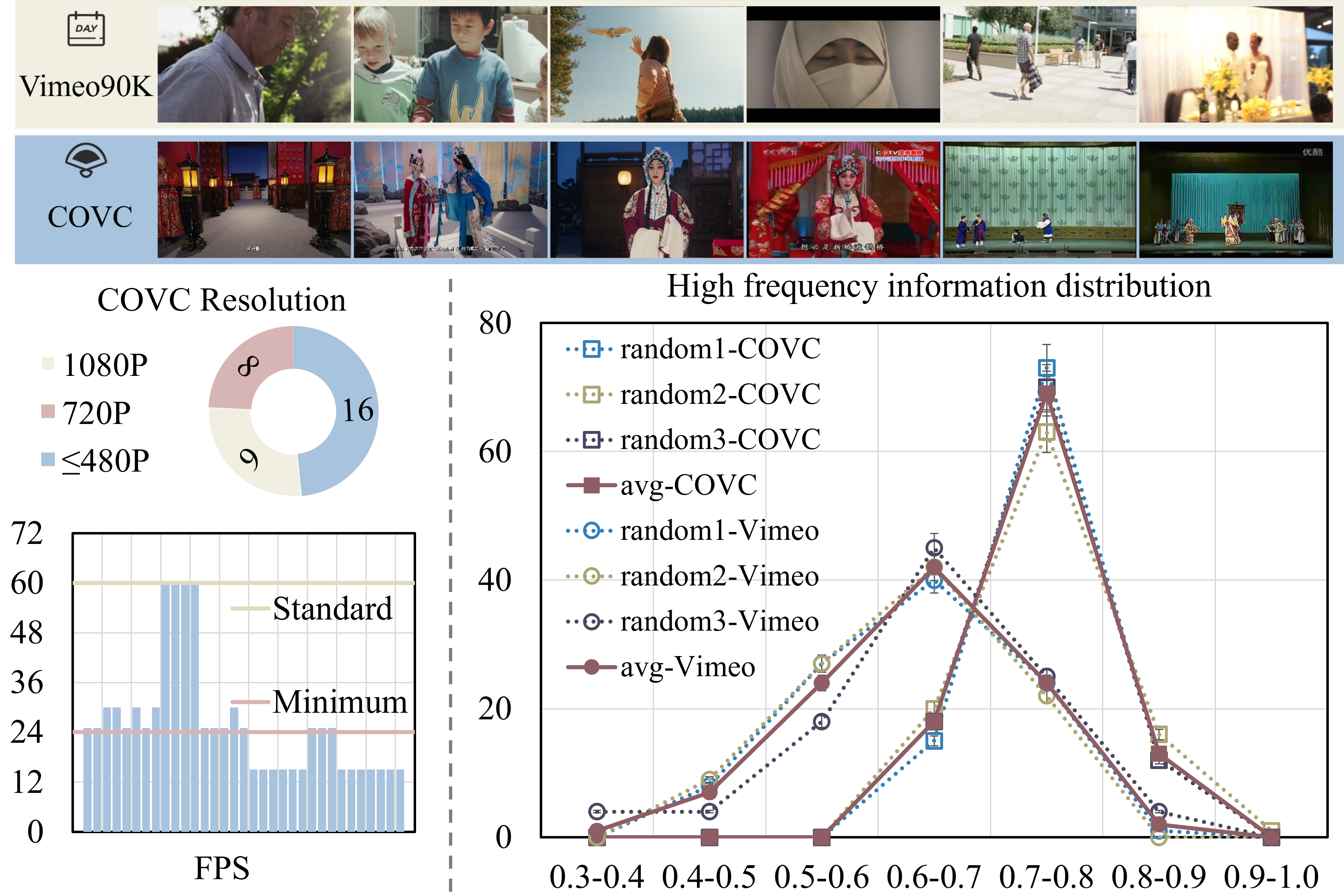}\vspace*{-2mm}
   \caption{Comparison of COVC and Vimeo samples and statistical data of COVC. Please zoom in for the best view.}
   \label{dataset}\vspace*{-4mm}
\end{figure}
\begin{figure*}[t]
  \centering
   \includegraphics[width=0.95\linewidth]{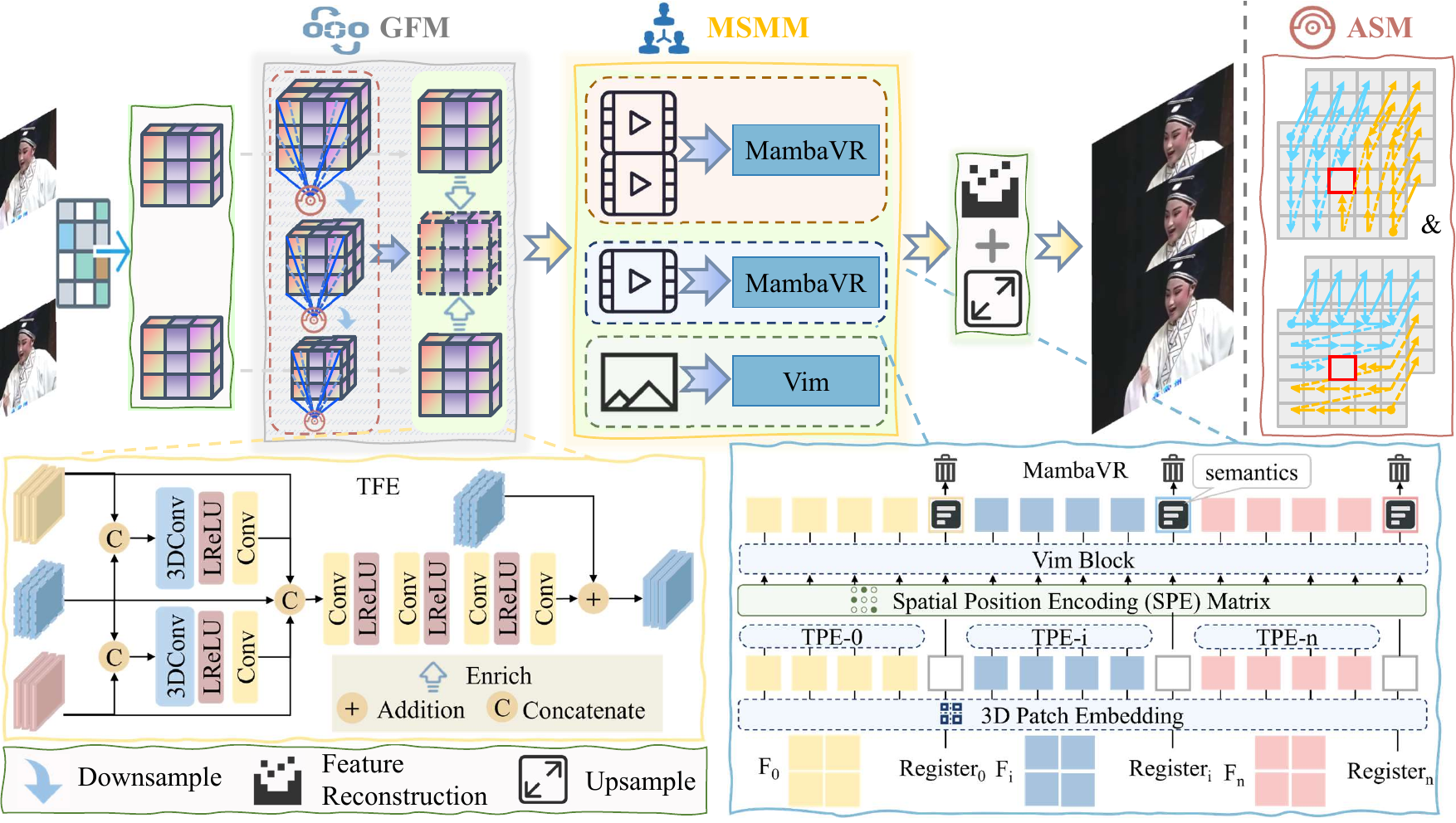}
   \caption{Architecture of the proposed Mamba-Based multiscale fusion network. Firstly, the features are extracted, and the missing intermediate frame features are obtained by the Global Fusion Module (GFM) with a multiscale alternating scanning mechanism (MASM). Next, each frame feature is enhanced by aligning sequences of different lengths using the Multiscale Synergistic Mamba Module (MSMM). Finally, high-quality video is obtained by feature reconstruction and PixelShuffle.}
   \label{framework}\vspace*{-4mm}
\end{figure*}

The COVC dataset comprises 33 opera videos: 9 at 1080p, 8 at 720p, and 16 at $\leq$480p (see \Cref{dataset}). Frame rates cluster predominantly around 24 fps, satisfying the minimum cinematic standard \cite{tag2016eye}, while four clips meet the 60 fps standard \cite{mackin2017investigating}. The top panel of \Cref{dataset} presents representative frames from COVC and Vimeo90K \cite{xue2019video}. While Vimeo90K videos primarily depict general scenes, opera clips—with their elaborate makeup, costumes, and stage settings—exhibit richer high‑frequency textures. This difference is highlighted by the line chart in \Cref{dataset} (bottom-right). Since COVC contains 1.6× more clips than Vimeo90K, we randomly sampled three sets of 100-frame sequences from both to quantify this difference. For each set, we computed the per‑frame proportion of high‑frequency information and plotted the results, along with the three‑trial average. Vimeo90K frames exhibit approximately 65\% high‑frequency content, whereas COVC frames concentrate around 75\%, with even the minimum exceeding 55\%. These findings demonstrate that Chinese opera videos possess richer texture complexity. Furthermore, since existing methods cannot model the large motions in opera videos (see \Cref{issue}(b)), we propose the following Mamba-based multiscale fusion network.

\subsection{Network Overview}
\label{network overview}
\noindent The proposed Mamba-Based multiscale fusion network, shown in \Cref{framework}, which aims to obtain high-resolution (HR), high frame rate (HFR) opera video frames $I^H=\left\{ I_{t}^{H} \right\} _{t=1}^{2n+1}$ with dimensions $3\times nH\times nW$, where n is the spatial upsampling factor, using low-resolution (LR), low frame rate (LFR) video sequence $I^L=\left\{ I_{2t-1}^{L} \right\} _{t=1}^{n+1}$. First, the feature extraction module, which includes a convolution layer and five residual modules, extracts input frame features $F^L=\left\{ F_{2t-1}^{L} \right\} _{t=1}^{n+1}$. These features are then passed to the Global Fusion Module (GFM), which generates missing intermediate frame features $F^L=\left\{ F_{2t}^{L} \right\} _{t=1}^{n}$. Next, the complete sequence is input into the Multiscale Synergistic Mamba Module (MSMM) to obtain enhanced high-quality (HQ) features $F^H=\left\{ F_{t}^{H} \right\} _{t=1}^{2n+1}$. Finally, the feature reconstruction and Pixelshuffle \cite{shi2016real} modules generate the HQ sequential frames $I^H=\left\{ I_{t}^{H} \right\} _{t=1}^{2n+1}$.

\subsection{Global Fusion Module}
\label{subsec: Global Fusion Module}
\noindent Deformable Convolution (DConv) \cite{dai2017deformable} enables efficient alignment by dynamically sampling spatial locations in a feature map. ZSM \cite{xiang2020zooming} first leveraged DConv for synthesizing missing intermediate frames, yielding promising results, and subsequent methods have widely adopted DConv for this purpose \cite{xu2021temporal,hu2023cycmunet+,wang2023stdan}. However, the fixed kernel size of DConv limits its receptive field, degrading performance on sequences with large motions. More recently, Mamba \cite{gu2023mamba} has emerged as an efficient global modeling framework in computer vision; VFIMamba \cite{zhang2024vfimamba} applied Mamba to VFI, but it still falls short in capturing fine-grained motion variations.

\textbf{Global Fusion Module.} Inspired by this, we propose the Global Fusion Module (GFM), which employs a Multiscale Alternating Scanning Mechanism (MASM) to globally model adjacent frame features and accurately capture large inter-frame motions. To synthesize intermediate frames, we fuse predictions from both forward and backward directions, wherein each direction learns global-to-local motion offsets via a multiscale pyramid architecture. Taking the forward synthesis direction ($0\rightarrow t$) as an example, we first downsample the neighboring frame features to multiple scales. At each scale, we merge the preceding and succeeding frame features into a single large feature map and globally arrange corresponding pixels along four directions (as shown in \Cref{framework}, top-right). We then compute the motion offset $H^{(N)}_{i,0\rightarrow t}$ by modeling pixel displacements between $F_{i-1}^L$ and $F_{i+1}^L$, and fuse this offset with the succeeding frame features to obtain the predicted intermediate frame at the current scale, $F_{i,0\rightarrow t}^{\left( N \right)}$. Finally, both the small-scale offset and its predicted intermediate frame are upsampled and integrated with larger-scale predictions to yield the final multiscale fusion result, $F^{L}_{i,0\rightarrow t}$. 
\begin{equation}
H^{(N)}_{i,0\rightarrow t}
\;=\;
MASM^{(N)}_{0\rightarrow t}\bigl(\downarrow^{(N)}(F_{i-1}^L),\;\downarrow^{(N)}(F_{i+1}^L)\bigr),
\end{equation}
\begin{equation}
F_{i,0\rightarrow t}^{\left( N \right)}=Fuse\left( H^{(N)}_{i,0\rightarrow t},\downarrow^{(N)}(F_{i+1}^L) \right), 
\end{equation}
\begin{equation}
F^{L}_{i,0\rightarrow t}
= Fuse\bigl(
    \uparrow^{(N\to N-1)}\bigl(F^{(N)}_{i,0\rightarrow t}\bigr),\;
    F^{(N-1)}_{i,0\rightarrow t}
  \bigr),
\end{equation}
where $i-1$, $i$, $i+1$ denote three consecutive frames; $N$ is the number of layers in the multiscale pyramid. $\downarrow$ indicate down-sampling and $\uparrow$ indicate up-sampling.

Finally, we fuse the forward and backward predictions to produce the final intermediate frame features. While the GFM module generates a complete frame sequence, the initially synthesized intermediate frames may exhibit minor artifacts under large-motion conditions.

\textbf{Temporal Feature Enhancement.} 
To refine the initially synthesized intermediate-frame features, we introduce the Temporal Feature Enhancement (TFE) module (see \Cref{framework}, bottom‐left). TFE concatenates the preceding, current, and succeeding frame features and processes them through a 3DConv–ReLU block to extract bidirectional motion offsets. These offsets are then concatenated with the original frame trio and passed through a multi‐layer convolutional blending network to yield refined intermediate features. Finally, we add the original intermediate features to the refined output to produce the enhanced frame representation. By leveraging adjacent frames for local convolutional refinement, TFE recovers fine-grained details and improves the alignment information available to subsequent modules.

\subsection{Multiscale Synergistic Mamba Module}
\label{subsec: Multiscale Synergistic Mamba}
\noindent Using the GFM, we generate video sequence features at high frame rates. Most of the existing methods use either the pass-through or sliding window approach for global frame sequence alignment \cite{xiang2020zooming,xu2021temporal,wang2023stdan}. The pass-through method accumulates alignment errors, which gradually affect subsequent frames as they are passed through, while sliding‐window approaches are confined to a fixed temporal neighborhood and cannot capture long‐range dependencies. Although 3D convolution can achieve global alignment by concatenating multiple frames \cite{fu20243dattgan}, it is only applicable to short-sequence videos and captures limited temporal information. VideoMamba \cite{li2025videomamba} has attracted attention as a potential alternative, but we found that its inherent feature artifacts and fixed-position encoding make it unsuitable.

\textbf{Mamba for Video Restoration (MambaVR).}
To address these challenges, we propose the MambaVR block specifically for video restoration (see \Cref{framework}, bottom‐right). First, in each frame’s feature map, we uniformly insert a fixed number of blank register tokens to buffer high‑norm semantic activations that could introduce feature artifacts. Video restoration demands strict local consistency—unlike Mamba‑R’s \cite{wang2024mamba} use of VisionMamba \cite{zhu2024vision} to encode high‑level semantics into background regions for classification, MambaVR isolates those semantics in removable tokens and discards them during reconstruction to preserve fine‑grained structure.

Second, we introduce Flexible Rotary Position Embedding (F‑RoPE) to overcome VisionMamba’s fixed, depth‑attenuating embeddings \cite{zhu2024vision}. F‑RoPE extends RoPE \cite{su2024roformer} by generating relative spatial encodings on‑the‑fly for any input resolution. It constructs base frequency tensors for the input dimensions ($D\times H\times W$), transforms them into a Spatial Position Encoding (SPE) matrix, and injects precise positional cues into the self‑attention mechanism via element‑wise multiplication.
\begin{equation}
  \omega_i = [\pi \cdot \frac{i}{2} ], \quad i = 1, 2, \dots, \frac{D}{2},
  \label{eq:4}
\end{equation}
\begin{equation}
\mathbf{f}_h(u) = [\,u\,\omega_i\,]_{i=1}^{D/2},\;
\mathbf{f}_w(v) = [\,v\,\omega_i\,]_{i=1}^{D/2},
\end{equation}
\begin{equation}
\mathrm{SPE}(u,v) = \mathrm{broadcat}\bigl(\mathbf{f}_h(u),\;\mathbf{f}_w(v)\bigr) \in \mathbb{R}^{D},
\end{equation}
where $u=0,\dots,H-1,\;v=0,\dots,W-1$. $broadcat$ is the original broadcasting mechanism.

\begin{figure*}[!t]
  \centering
   \includegraphics[width=1\linewidth]{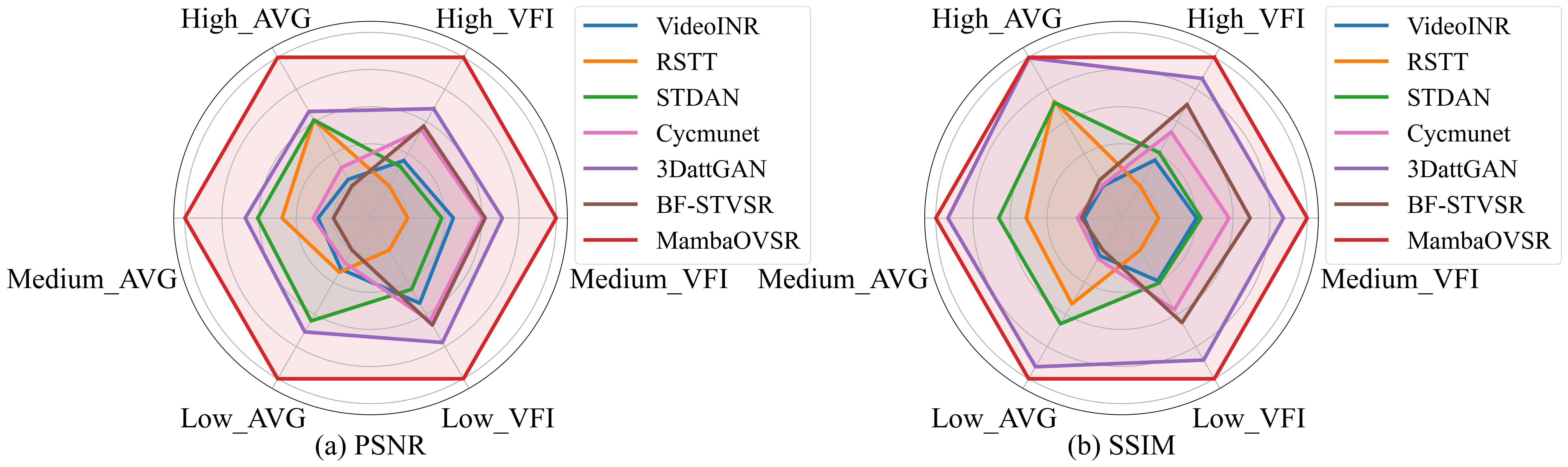}\vspace*{-2mm}
   \caption{Quantitative comparison with the 
  Other Space-Time Video Super-Resolution (STVSR) methods on COVC. (a) depicts a radar plot for PSNR comparisons between all generated frames (AVG) and for interpolated frames (VFI) on the three test sets, High, Medium and Low, while (b) depicts a radar plot for SSIM. Note that all metrics have been normalized, and detailed metric results can be found in Table 3 of Appendix Section C.1.}
   \label{Quantitative comparison}\vspace*{-4mm}
\end{figure*}

\textbf{Multiscale Synergistic Mamba Module.}
To achieve multi-granularity motion alignment, we propose the Multiscale Synergistic Mamba Module (MSMM), built upon our MambaVR block and Vim \cite{zhu2024vision} (see \Cref{framework}). \textbf{Global implicit alignment:} Feed the full sequence into MambaVR for holistic feature interaction. \textbf{Short-term temporal consistency:} Apply a sliding window over segments to preserve local motion coherence. \textbf{Global guidance enhancement:} Use MambaVR’s hidden state to update Vim’s, enriching each frame with global context. As an example of global alignment, the full sequence features are concatenated and passed through a 3D convolution to generate successive temporal patches (of length $L$). Then, we uniformly insert $n$ register tokens ($r$) into the sequence, and the temporal position encoding (TPE) $P_t\in R^{T\times C}$ is added.
\begin{equation}
  X\ =\ \left[ \cdots x_i,r_1,\cdots x_{2i},r_2,\cdots x_{ni},r_n,\cdots x_L \right] +P_t,  
  \label{eq:9}
\end{equation}
where $T$ denotes the sequence length, $C$ the channel dimension. Next, we apply element‑wise multiplication with the Spatial Position Encoding $\mathrm{SPE}(u,v)$ and feed the result into the MambaVR block to obtain globally aligned features $E_g$:
\begin{equation}
  X\ = X \otimes \mathrm{SPE}(u,v) ,\ E_g=\mathrm{MambaVR}\left( X \right),  
  \label{eq:10}
\end{equation}
To preserve short-term consistency, we feed three consecutive frames sequentially into distinct MambaVR blocks, yielding short‑term aligned sequences $E_{l}^{j}$. Additionally, we initialize Vim’s hidden state with that of the global MambaVR, thereby leveraging global context to guide per-frame feature enhancement.
\begin{equation}
  L_{1}^{'},\cdots ,L_{l}^{'}\ =\ vim\left( F_{1}^{L},\cdots ,F_{l}^{L} \right).  
  \label{eq:11}
\end{equation}
Frame‐specific supplementary information is obtained by concatenating features across multiple scales:
\begin{equation}
  F_{i}^{E}\ =\ concat\left( E_g,E_{l}^{j},L_{i}^{'} \right),
  \label{eq:12}
\end{equation}
where $i$ indexes the current frame and $j$ denotes distinct short-term contexts.
Leveraging residual connections, we integrate MSMM‑extracted features—refined via channel attention and projected back to the original dimensionality through a $1\times 1$ convolution—with the original frame features preserved by an initial convolution.
\begin{equation}
  F_{i}^{H}=conv\left( F_{i}^{L} \right) +conv1D\left( attn_i\left( F_{i}^{E} \right) \right), 
  \label{eq:13}
\end{equation}
where $attn$ is channel attention and $conv$ a convolution layer; see \textbf{Appendix E} for the frame reconstruction module.

\section{Experiment}
\begin{figure*}[!t]
  \centering
   \includegraphics[width=0.9\linewidth]{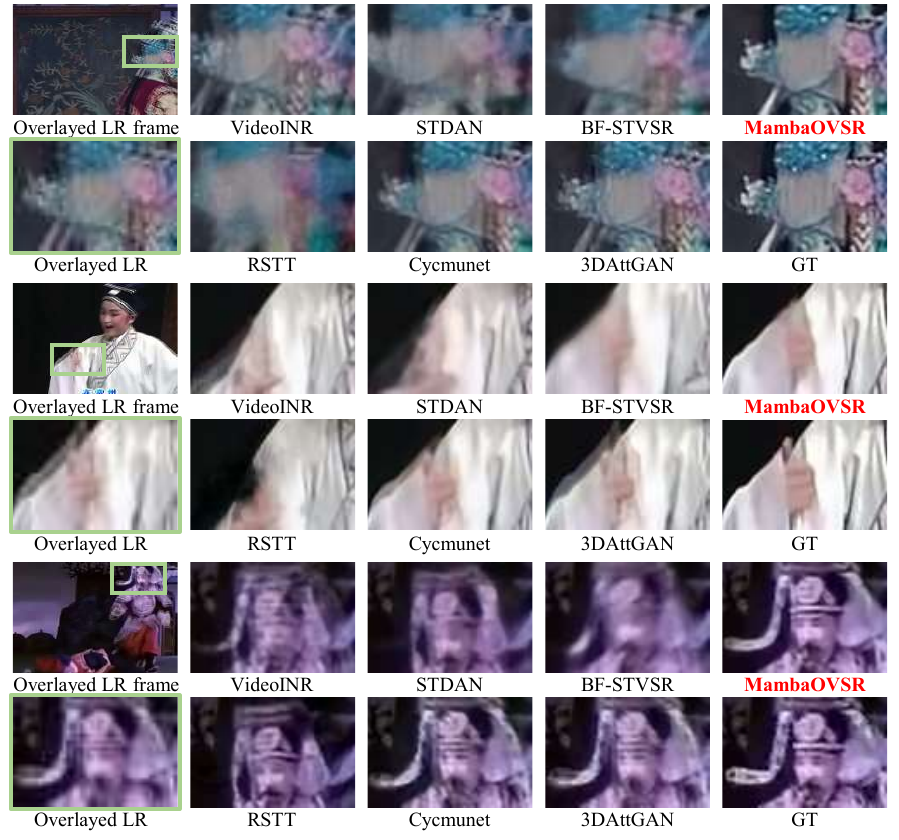}\vspace*{-2mm}
   \caption{Qualitative Comparisons of the different approaches on three qualities of Chinese opera videos, from top to bottom, for the High, Medium and Low test sets. Our framework can recover more details while producing fewer artifacts.}
   \label{visual comparison}\vspace*{-4mm}
\end{figure*}

\noindent \textbf{Implement Details.}
Even‑indexed frames are downsampled 2× for input, with the seven‑frame sequence as supervision. Frames are randomly cropped to 128×128, downscaled to 64×64, and augmented via flips and rotations. We train with batch size 8, using an initial learning rate of 0.01 decayed to $1\times10^{-7}$ via cosine annealing \cite{gotmare2018closer}, and optimize with AdaMax ($\beta_1=0.9$ and $\beta_2=0.999$). All experiments are implemented in PyTorch 2.1.

\noindent \textbf{Datasets and Metrics.}
We retrained other STVSR methods on the introduced COVC and general Vimeo90K \cite{xue2019video} dataset and quantitatively evaluated the performance of the different models using Peak Signal-to-Noise Ratio (PSNR) and Structural Similarity Index (SSIM) \cite{wang2004image} as evaluation metrics.

\begin{table}[!thbp]
  \centering
  \setlength{\tabcolsep}{50  mm}
  \begin{tblr}{
    width = \textwidth,
    colspec = {c||c|cc},  
    hline{1} = {1pt},
    hline{2,3,4,5,6,7} = {0.5pt}, 
    hline{Z} = {1pt},
    column{1} = {font=\bfseries},
  }
  Methods&Venue& PSNR$\uparrow$ &SSIM$\uparrow$\\
  \midrule
  VideoINR &CVPR'22&20.41&0.6518\\
  RSTT &CVPR'22&29.09&0.7996\\
  Cycmunet &TPAMI'23&21.30&0.6532\\
  3DAttGAN &TETCI'24&\underline{30.65}&\underline{0.8371}\\
  BF-STVSR &CVPR'25&20.67&0.6762\\
  \SetCell[r=2]{c}MambaOVSR&\SetCell[r=2]{c}-&\textbf{35.61} &\textbf{0.8794}\\
  &&$\uparrow$4.96&$\uparrow$0.0423\\
  \end{tblr}

  \caption{Quantitative comparison on Vimeo90K Fast subset.}
  \label{Quantitative comparison on Vimeo90K fast}\vspace*{-4mm}
\end{table}

\subsection{Comparison of Methods}
\noindent We present a comprehensive comparison of our framework with existing STVSR methods, including VideoINR \cite{chen2022videoinr}, RSTT \cite{geng2022rstt}, STDAN \cite{wang2023stdan}, Cycmunet \cite{hu2023cycmunet+}, 3DAttGAN \cite{fu20243dattgan}, and BF-STVSR \cite{kim2025bf}. To ensure a fair comparison, we retrained these models on the COVC.

As shown in \Cref{Quantitative comparison}, we present radar plots of PSNR (a) and SSIM (b) for all generated frames (AVG) and interpolated frames (VFI) across the High, Medium, and Low test subsets. MambaOVSR achieves significant improvements in both metrics—particularly PSNR, with relative gains of 6.51\%, 6.24\%, and 5.24\% over the SOTA 3DAttGAN on the three subsets. These results confirm the method’s effectiveness in modeling large motions. Detailed metric results are provided \textbf{Table 3 of Appendix section C.1}.

The visual comparison of the methods is presented in \Cref{visual comparison}. Existing approaches struggle to handle large motion, leading to pronounced blurring artifacts in the synthesized frames. In contrast, our method produces far fewer blurs and recovers finer details, further demonstrating its effectiveness. Moreover, it achieves these results while maintaining moderate computational complexity; detailed comparisons are provided in \textbf{Table 1 of Appendix section C.2}.


Furthermore, to validate MambaOVSR’s capability in modeling large motions, we compared several methods on the Vimeo90K Fast test set, which is characterized by large motions. As shown in \Cref{Quantitative comparison on Vimeo90K fast}, MambaOVSR achieved SOTA. Full results are shown in \textbf{Appendix section C.3}.

\subsection{Ablation Studies}

\begin{figure}[!t]
  \centering
   \includegraphics[width=1\linewidth]{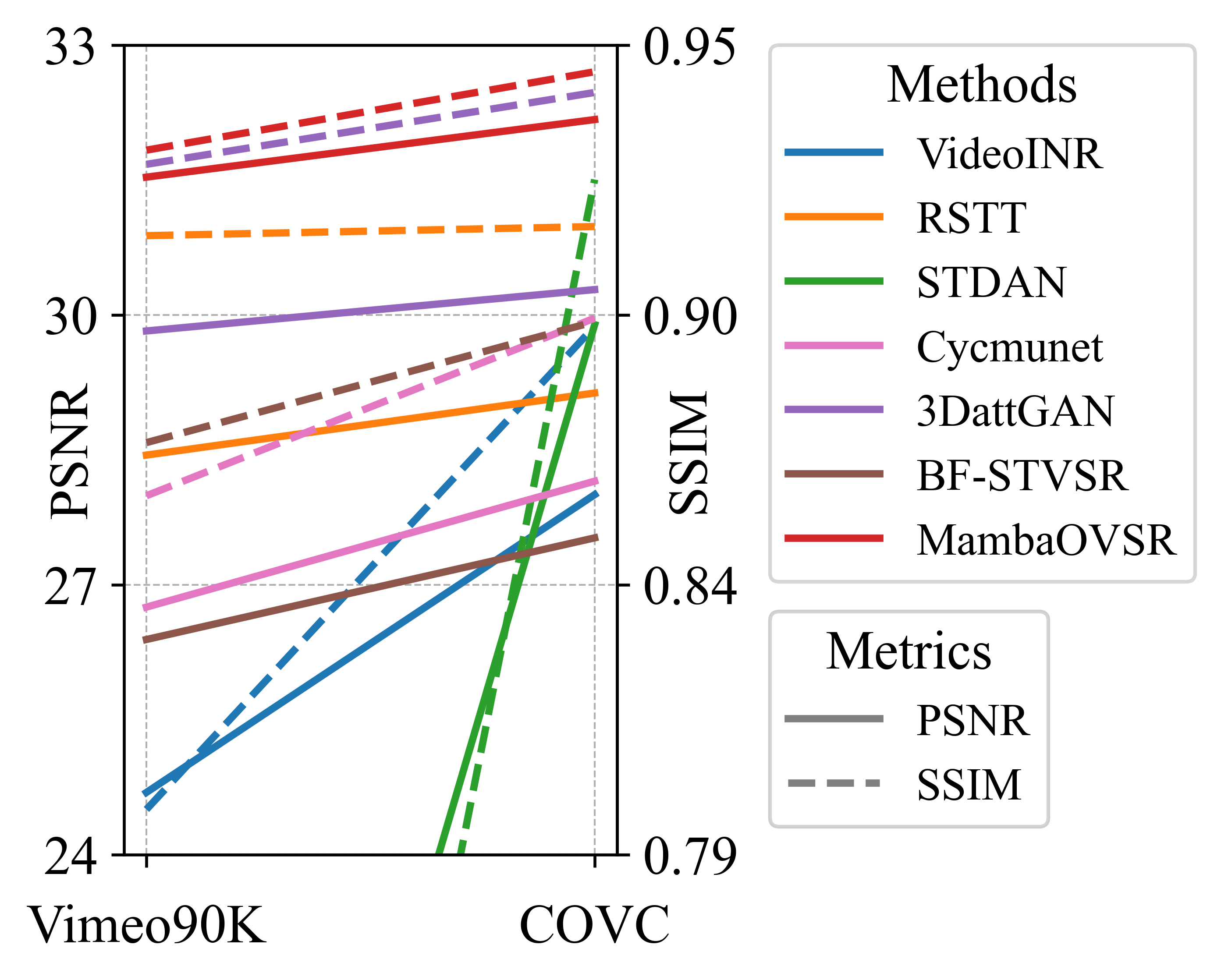}\vspace*{-2mm}
   \caption{Quantitative comparison of methods trained on the Vimeo90K and COVC datasets.}
   \label{Quantitative comparison1}
\end{figure}

\noindent To validate the effectiveness of each proposed module, we further conducted the following comprehensive ablation studies on the Medium test set.
\begin{table}[!t]
  \centering
  \begin{tblr}{
    width = \linewidth,
    colspec = {c||ccccc},  
    hline{1} = {1pt},
    hline{2} = {0.5pt},
    hline{4} = {0.5pt},
    hline{Z} = {1pt},
    vline{3-6} = {0.5pt}, 
    column{1} = {font=\bfseries},
  }

  Methods&$\varOmega_1$ & $\varOmega_2$& $\varOmega_3$& $\varOmega_4$& $\varOmega_5$ \\
  \midrule
  DConv& \usym{2713}&\usym{2717} &\usym{2713} &\usym{2713} &\usym{2717} \\
  \textbf{GFM}&\usym{2717} &\usym{2713} &\usym{2717} &\usym{2717} &\usym{2713} \\
  Cyc&\usym{2713} &\usym{2713} &\usym{2717} &\usym{2717} &\usym{2717} \\
  MSTM&\usym{2717} &\usym{2717} &\usym{2713} &\usym{2717} &\usym{2717} \\
  \textbf{MSMM}&\usym{2717} &\usym{2717} &\usym{2717} & \usym{2713}&\usym{2713} \\
  \midrule
  AVG&28.15 &28.37 &31.67 &31.86 &\textbf{32.17} \\
  VFI&28.00 &28.31 &30.53 &30.74 &\textbf{31.05} \\
  
  \end{tblr}
  \caption{Ablation study results for GFM and MSMM are presented via PSNR comparisons. Cyc denotes the baseline, while MSTM refers to the MSMM module incorporating a Motionformer transformer block.}\vspace*{-3mm}
  \label{Ablation}
\end{table}

\noindent \textbf{Effectiveness of COVC.}
To validate the effectiveness of the COVC dataset, we trained all comparison methods on Vimeo90K and COVC under the same configuration, and compared the AVG performance on the Medium test set (see \Cref{Quantitative comparison1}). The results show that the model trained on COVC consistently outperforms Vimeo90K, both in PSNR and SSIM, corresponding to the upward trend of the lines in the figure. Meanwhile, MambaOVSR performs the best among all methods, with the line at the top, proving its excellent generalization ability. The full metrics results are detailed in \textbf{Table 5 of Appendix section C.4}.

\noindent \textbf{Effectiveness of MSMM.}
To assess the effectiveness of the proposed MSMM, we designed three models: $\varOmega_1$, $\varOmega_3$, and $\varOmega_4$. Each model utilizes a deformable convolution-based module for intermediate feature interpolation. $\varOmega_1$ leverages space-time correlation through up-and-down projections \cite{hu2023cycmunet+}, while $\varOmega_4$ uses the MSMM module to align sequences of varying lengths implicitly. To evaluate the effectiveness and efficiency of the Mamba framework in video modeling, we replace the MambaVR and Vim blocks of $\varOmega_3$ with Motionformer \cite{patrick2021keeping}. 

\begin{figure}[!t]
  \centering
   \includegraphics[width=1\linewidth]{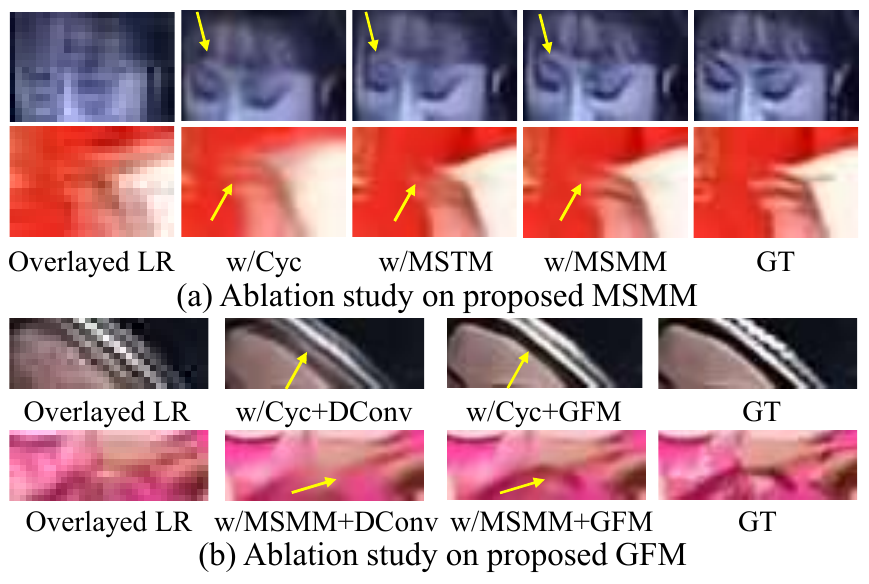}\vspace*{-2mm}
   \caption{Qualitative Comparisons of proposed GFM and MSMM for ablation studies. DConv denotes the deformable convolution-based module.}
   \label{ablation1}
\end{figure}

\Cref{Ablation} shows that both $\varOmega_3$ and $\varOmega_4$ significantly outperform $\varOmega_1$ in terms of PSNR and SSIM. Integrating our MambaVR block yields improvements of 3.71 dB for AVG and 2.74 dB for VFI. A comparison of visual effects is shown in \Cref{ablation1} (a), where both $\varOmega_3$ and $\varOmega_4$ exhibit better clarity and detail richness than the baseline model $\varOmega_1$, while the MSTM shows slight blurring of edge structure compared to MSMM. The full quantitative and qualitative comparisons are provided in \textbf{Appendix section D}.

\noindent \textbf{Effectiveness of GFM.}
To verify the effectiveness of the proposed GFM, we constructed two parallel ablation studies on distinct baseline architectures: $\left( \varOmega_1, \varOmega_2 \right) $ and $\left( \varOmega_4, \varOmega_5 \right) $. Specifically, $\varOmega_1$ and $\varOmega_4$ each use the deformable convolution (DConv) based feature interpolation module, while $\varOmega_2$ and $\varOmega_5$ replace it with the GFM. Quantitative results (see \Cref{Ablation}) show that integrating GFM consistently yields a PSNR improvement of approximately 0.2–0.3 dB over the DConv counterparts across both baseline networks. Visual comparisons in \Cref{ablation1} (b) further confirm that GFM yields sharper edges and more accurate fine structures. These studies demonstrate that GFM reliably enhances detail recovery and reconstruction fidelity in video super‐resolution.

\begin{figure}[!t]
  \centering
   \includegraphics[width=1\linewidth]{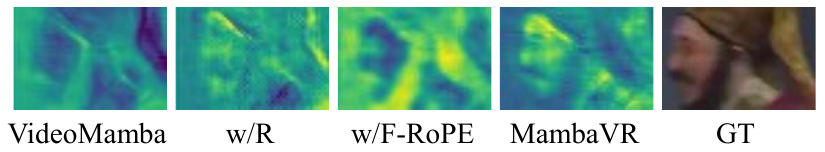}\vspace*{-2mm}
   \caption{Feature map of MambaVR ablation study. 
   }
   \label{ablation_MambaVR}\vspace*{-4mm}
\end{figure}

\noindent \textbf{Effectiveness of MambaVR.}
The ablation study results for our MambaVR block are reported in \Cref{Ablation_MambaVR_table}. It can be seen that all MambaVR variants outperform the vanilla VideoMamba block \cite{li2025videomamba} in both PSNR and SSIM. Crucially, Registers and F-RoPE act synergistically: their combination yields the greatest improvement in reconstruction quality. As shown in the feature‐map visualizations in \Cref{ablation_MambaVR}, VideoMamba alone generates overly blurred facial regions; incorporating Registers noticeably reduces this blur, while F-RoPE further sharpens facial contours. By integrating both Registers and F-RoPE, MambaVR harnesses their complementary strengths, producing the most detailed and accurate facial reconstructions.

\begin{table}[!t]
  \centering
  \begin{tblr}{
    width = \linewidth,
    colspec = {c||cccc},  
    hline{1} = {1pt},
    hline{2,3} = {0.5pt},
    hline{6} = {0.5pt},
    hline{Z} = {1pt},
    vline{4} = {0.5pt}, 
    row{1,2} = {font=\bfseries}, 
  }

  \SetCell[r=2]{c} Methods & \SetCell[c=2]{c} PSNR & & \SetCell[c=2]{c} SSIM &\\
  &AVG & VFI & AVG & VFI \\
  \midrule
  VideoMamba &31.72 &30.65 &0.9368 &0.9259 \\
  w/R &31.82 &\textbf{30.79} &0.9422 &0.9328\\
  w/F-RoPE &31.80 &30.71 &0.9419 &0.9320\\
  MambaVR &\textbf{31.86} &30.74 &\textbf{0.9438} &\textbf{0.9336}\\
  
  \end{tblr}
  \caption{Ablation study on the proposed MambaVR. w/ denotes inclusion of each enhancement.}
  \label{Ablation_MambaVR_table}
  
\end{table}

\section{Conclusion}


\noindent In this work, we built a large-scale Chinese opera video clip (COVC) dataset and introduced the Mamba-Based multiscale fusion network for space-time Opera Video Super-Resolution (MambaOVSR). Specifically, we designed a global fusion module (GFM) for fine-grained holistic motion modeling between adjacent frames. Additionally, we proposed a MambaVR block to achieve global alignment. Based on this, our Multiscale Synergistic Mamba Module (MSMM) implemented granular motion alignment across varying sequence lengths. Experimental results on the COVC and Vimeo90K dataset showed that our method significantly outperforms existing STVSR methods. Future work will focus on optimising computational efficiency.

\bibliography{aaai2026}

\end{document}